\documentclass{article}

\PassOptionsToPackage{numbers, compress}{natbib}
\usepackage[final]{cpal_2024}

\usepackage[utf8]{inputenc} 
\usepackage[T1]{fontenc}    
\usepackage{hyperref}       
\usepackage{url}            
\usepackage{booktabs}       
\usepackage{amsfonts}       
\usepackage{nicefrac}       
\usepackage{microtype}      
\usepackage{xcolor}         
\usepackage{amsmath}
\usepackage{graphicx}
\usepackage{subcaption}
\usepackage{caption}
\usepackage{listings}
\usepackage{xspace}

\lstdefinestyle{promptblock}{
    basicstyle=\ttfamily\small,
    backgroundcolor=\color{gray!10},
    frame=single,
    breaklines=true,
    captionpos=b,
    xleftmargin=0pt,
    xrightmargin=0pt
}

\newcommand\map[2]{#1\rightarrow#2}

\newcommand{\pmi}{\text{pmi}_{\sD}(c_1,c_2)}
\newcommand{\pmifunc}{\text{pmi}_{\sD}}

\newcommand{\pairprob}{p_{\sD}(c_1, c_2)}
\newcommand{\singleprob}{p_{\sD}(c)}
\newcommand{\conceptdist}{p_{\sD}}
\newcommand{\otherword}{c_{\text{accessory}}}
\newcommand{\imgnetword}{c_{\text{ImageNet}}}
\newcommand{\synth}{\text{GenPairs}\xspace}
\newcommand{\edited}{\text{ImageNet-Paste}\xspace}

\newcommand{\indicator}{\mathbf{1}}

\DeclareMathOperator*{\argmax}{arg\,max}

\newlength{\widebarargwidth}
\newlength{\widebarargheight}
\newlength{\widebarargdepth}

\newcommand\sC{\ensuremath{\mathcal{C}}}
\newcommand\sD{\ensuremath{\mathcal{D}}}

\newcommand\sT{\ensuremath{\mathcal{T}}}

\newcommand\sX{\ensuremath{\mathcal{X}}}

\newcommand\sZ{\ensuremath{\mathcal{Z}}}

\newcommand\BR{\ensuremath{\mathbb{R}}}

\newcommand\eqnl[2]{\begin{align} \label{eqn:#1} #2 \end{align}} 

\ifthenelse{\isundefined{\definition}}{\newtheorem{definition}{Definition}}{}
\ifthenelse{\isundefined{\assumption}}{\newtheorem{assumption}{Assumption}}{}
\ifthenelse{\isundefined{\hypothesis}}{\newtheorem{hypothesis}{Hypothesis}}{}
\ifthenelse{\isundefined{\proposition}}{\newtheorem{proposition}{Proposition}}{}
\ifthenelse{\isundefined{\theorem}}{\newtheorem{theorem}{Theorem}}{}
\ifthenelse{\isundefined{\lemma}}{\newtheorem{lemma}{Lemma}}{}
\ifthenelse{\isundefined{\corollary}}{\newtheorem{corollary}{Corollary}}{}
\ifthenelse{\isundefined{\alg}}{\newtheorem{alg}{Algorithm}}{}
\ifthenelse{\isundefined{\example}}{\newtheorem{example}{Example}}{}

\def\shownotes{1}  
\ifnum\shownotes=1
\newcommand{\authnote}[2]{[#1: #2]}
\else
\newcommand{\authnote}[2]{}
\fi

\title{Impact of Pretraining Word Co-occurrence on Compositional Generalization in Multimodal Models}

\author{%
  Helen Qu\\
  Flatiron Institute\\
  \texttt{hqu@flatironinstitute.org} \\
  \And
  Sang Michael Xie \\
  Stanford University\\
  \texttt{xie@cs.stanford.edu} \\
}

\begin{document}

\maketitle

\begin{abstract}
CLIP and large multimodal models (LMMs) have better accuracy on examples involving concepts that are highly represented in the training data. 
However, the role of concept \emph{combinations} in the training data on compositional generalization is largely unclear -- for instance, how does accuracy vary when a common object appears in an uncommon pairing with another object?
In this paper, we investigate how word co-occurrence statistics in the pretraining dataset (a proxy for co-occurrence of visual concepts) impacts CLIP/LMM performance.
To disentangle the effects of word co-occurrence frequencies from single-word frequencies, we measure co-occurrence with pointwise mutual information (PMI), which normalizes the joint probability of two words co-occurring by the probability of co-occurring independently.
Using synthetically generated images with a variety of concept pairs, we show a strong correlation between PMI in the CLIP pretraining data and zero-shot accuracy in CLIP models trained on LAION-400M ($r=0.97$ and 14\% accuracy gap between images in the top and bottom 5\% of PMI values), demonstrating that even accuracy on common concepts is affected by the combination of concepts in the image.
Leveraging this finding, we reproduce this effect in natural images by editing them to contain pairs with varying PMI, resulting in a correlation of $r=0.75$.
Finally, we demonstrate that this behavior in CLIP transfers to LMMs built on top of CLIP ($r=0.70$ for TextVQA, $r=0.62$ for VQAv2).
Our findings highlight the need for algorithms and architectures that improve compositional generalization in multimodal models without scaling the training data combinatorially.
Our code is available at \texttt{\url{https://github.com/helenqu/multimodal-pretraining-pmi}}.
\end{abstract}

\section{Introduction}
Contrastive image-text encoders such as CLIP \citep{radford2021clip,cherti2023reproducible} are a crucial component of large multimodal models (LMMs) \citep{achiam2023gpt,liu2023visual,deitke2024molmo,awadalla2023openflamingo} that can perform a diverse array of vision-language tasks.
CLIP’s strong zero-shot accuracy on challenging datasets, such as ImageNet-R and ObjectNet \citep{taori2020measuring,hendrycks2020many,barbu2019objectnet}, indicates that a large, diverse pretraining dataset may be sufficient for robust generalization~\citep{fang2022data}.

To understand this, recent work has shown that CLIP accuracy on examples with a particular visual concept correlates strongly with the frequency of that concept in the pretraining data distribution \citep{parashar2024neglected,udandarao2024no}. Thus, a natural way forward is to scale the pretraining dataset for sufficient coverage on all concepts.
However, extending this idea to concept \textit{combinations} suggests that we need combinatorially many pretraining examples to generalize to all concept combinations, which is prohibitively expensive.
In general, the role of concept combinations in pretraining data and its effects on accuracy in both CLIP and LMMs built on CLIP is underexplored.
Prior compositional generalization benchmarks typically only test combinations unseen during pretraining \citep{wiedemer2025pretraining,abbasi2024deciphering} or are generic evaluations that do not consider the role of pretraining data \citep{thrush2022winoground,ma2023crepe,hsieh2023sugarcrepe,wang2024sober}.

In this work, we address this gap by investigating CLIP and LMM accuracies through the lens of the co-occurrence rate of concept pairs in the pretraining dataset.
In particular, we focus on co-occurrence between words in the textual captions of CLIP pretraining examples as a proxy for visual concepts.
To decorrelate the pair frequency from single-concept frequencies (i.e., the individual concepts in low frequency pairs are often themselves low frequency), we calculate pointwise mutual information (PMI) \citep{church-hanks-1990-word} for all concept pairs, which measures the probability of concept co-occurrence normalized by the expected probability if the concepts were independent.

\begin{figure}
    \centering
    \includegraphics[width=1\linewidth]{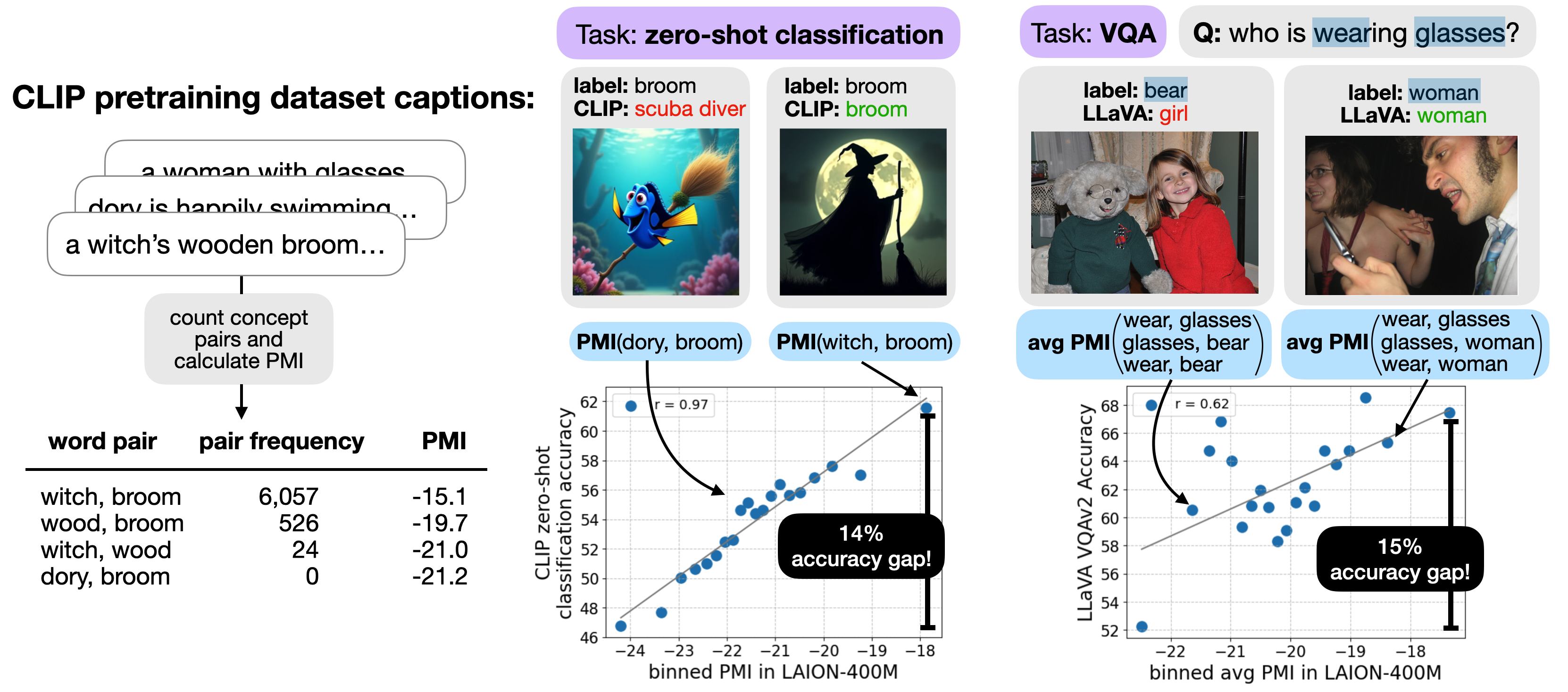}
    \caption{\textbf{Overview of our contributions. (left)}  We extract concept pairs from pretraining data caption text and calculate their co-occurrence frequency and pointwise mutual information (PMI) across all captions in the dataset, including pairs that do not co-occur in the dataset at all. \textbf{(middle)} We design a synthetic dataset with concept combinations across a wide range of PMI and find a strong correlation between CLIP zero-shot classification accuracy and PMI (i.e., the same concept ``broom'' is less accurately identified in a low vs. high PMI pair). \textbf{(right)} We continue to find an accuracy vs. PMI correlation in large multimodal models built on CLIP embeddings evaluated on visual question-answering tasks. In this case, per-example PMI is averaged over all the concept pairs in the question-answer pair (shown in blue highlights).}
    \label{fig:main}
\end{figure}

In our experiments, we find a strong correlation between concept pair PMI in the pretraining data and accuracy: in leading CLIP models trained on LAION-400M evaluated on zero-shot image classification, as well as in LMMs built on CLIP embeddings evaluated on visual question-answering.
We test this by generating synthetic images with a variety of concept pairs, finding a 14\% absolute (30\% relative) difference in zero-shot classification accuracy between images in the bottom vs. top 5\% of concept pair PMI.
With this understanding, we reproduce this correlation in natural images by simply pasting in a small image of an object that rarely co-occurs in pretraining with the main image object (low PMI), reducing CLIP zero-shot classification accuracy by up to 10\%.
Finally, we evaluate LLaVA, a leading LMM built on CLIP, and find that the correlation between PMI and task accuracy still holds on visual question-answering benchmarks.
This result even extends to LMMs built on closed-source embedding models such as OpenAI's CLIP.

Co-occurrence statistics in the pretraining data also lead to model biases.
In particular, we find that LLaVA exhibits an output bias that is correlated with co-occurrence, tending to output ``Yes'' for questions with highly co-occurring concepts regardless of the true label.
Overall, our findings show that CLIP and LMMs have biases that depend heavily on co-occurrence statistics from pretraining, rather than an understanding of the individual concepts.
Our results point to the need for new algorithms and architectures that improve compositional generalization capabilities in multimodal models.

\section{Setup}
We introduce the models and pretraining dataset used in our analyses.

\paragraph{CLIP.}
Contrastive Language-Image Pretraining (CLIP) \cite{radford2021clip} is a self-supervised learning method that uses natural language supervision in the form of image captions to learn downstream task-agnostic image representations.
Formally, in a batch of $N$ image-text pairs $\{(x_i, t_i)\}^N_{i=1}$ where $x_i \in \sX$, $t_i \sim \sT$, CLIP simultaneously trains an image encoder $\phi: \map{\sX}{\sZ_v}$ and text encoder $\psi: \map{\sT}{\sZ_t}$, where $\sZ_v \subset \BR^d$, $\sZ_t \subset \BR^d$ denote the image and text embedding spaces, respectively. 
The encoders are trained to minimize the multi-class $N$-pair loss:

\eqnl{clip_loss}{\ell_{\text{CLIP}}(\phi, \psi) =
- \frac{1}{N} \sum_i \ln \frac{e^{\phi(x_i)^\top \psi(t_i) / T}}{\sum_j e^{\phi(x_i)^\top \psi(t_j) / T}}
- \frac{1}{N} \sum_j \ln \frac{e^{\phi(x_j)^\top \psi(t_j) / T}}{\sum_i e^{\phi(x_i)^\top \psi(t_j) / T}}.
}

For a test image $x$, we can use the learned encoders for zero-shot classification by translating a list of classification label texts $y_1,\dots,y_k$ (where $k$ is the number of classes) into pseudo-captions $y’_1,\dots,y'_k$, e.g., $\texttt{a photo of \{class name\}}$, and selecting the class whose pseudo-caption embedding aligns best with the image embedding: $\argmax_i \phi(x)^\top \psi(y’_i)$.

\paragraph{LMMs.}
Large multimodal models (LMMs) synthesize data from multiple data modalities (e.g., image, text), typically building on top of a large language model (LLM) for natural language understanding.
Many state-of-the-art open-source LMMs (e.g., LLaVA \citep{liu2023visual,liu2024improved}, Molmo \citep{deitke2024molmo}) leverage a trained CLIP image encoder to compute visual features $\phi(x_i)$ from input image $x_i$.
A vision-language connector $h: \map{\sZ_v}{\sZ'_t}$ is trained to map these visual features into the language model’s embedding space $\sZ'_t$.
The language model and connector are then fine-tuned with conversational/question-answering data to optimize performance.

\paragraph{LAION-400M.}
LAION-400M \citep{schuhmann2021laion} is a dataset of 400 million image-text pairs curated from Common Crawl by filtering out pairs with CLIP embedding cosine similarity below 0.3. LAION-400M was created to emulate the closed-source WIT-400M~\citep{radford2021clip} dataset used to train the original CLIP implementation.

\section{Concept Pair Extraction and Quantifying Co-occurrence}

In this work, we study the impact of pretraining data on combinatorial generalization in CLIP and LMMs. To this end, we define a procedure for concept extraction from large image-text datasets as well as metrics to disentangle pair co-occurrence frequency from single concept frequency.

\paragraph{Concept and concept pair probability.}
We define the set of concepts $\sC$ as the set of lemmatized words extracted from a dataset of image captions $\sD$, where a concept $c \in \sC$ corresponds to a single lemmatized word.
A concept pair is an unordered pair of concepts $(c_1, c_2)$. 
We define the empirical distribution of single concepts $\singleprob$ as 
\eqnl{eqn:concept_freq}{
    \singleprob=\frac{1}{|\sC|} \sum_{d \in \sD}\indicator[c \in d]
}
where for simplicity we abuse the notation to define $d$ as the set of concepts derived from each caption in $\sD$.
Similarly, we define the probability of a concept pair $(c_1, c_2) \in \sC \times \sC$ as:
\eqnl{pair_freq}{
    \pairprob = \frac{1}{{|\sC| \choose 2}} \sum_{d\in \sD}\indicator[c_1 \in d \; \land \; c_2 \in d]
}

\paragraph{Pointwise Mutual Information (PMI).}
To decorrelate concept pair probabilities $\pairprob$ from their constituent single concept frequencies $\conceptdist(c_1), \conceptdist(c_2)$, we measure the pointwise mutual information (PMI) \citep{church-hanks-1990-word} between concept pairs:
\eqnl{pmi}{\pmi = \text{log}\Big( \frac{ \pairprob }{ \conceptdist(c_1) \conceptdist(c_2) } \Big)}
PMI measures how much more $c_1,c_2$ co-occur in $\sD$ than we would have expected them to appear by chance.
Note that while our analysis focuses on concept pairs, the PMI framework can be extended to any number of concepts $(c_1, c_2, ..., c_n)$ through specific correlation \citep{vandecruys2011two}:
\eqnl{eqn:si}{\text{si}_{\sD}(c_1, ..., c_n) = \text{log} \Big( \frac{\conceptdist(c_1, ..., c_n)}{\prod_{i=1}^{n}{\conceptdist(c_i)}} \Big) }

\paragraph{Concept Extraction and PMI Calculation in LAION-400M.}
While visual concepts can be difficult to define and extract from images, we leverage the textual captions from LAION-400M as a proxy for the visual concepts present in each image.
Starting with the set of LAION-400M captions, we remove stopwords and lemmatize each word.
In order to minimize noise in our metric, we filter captions to include only words with frequency greater than 10,000.
The remaining 21,718 unique words make up our concept set.
To calculate PMI, we count individual frequencies as well as pair frequencies for all concepts in the set.

\paragraph{Evaluation metrics.}
We introduce some metrics we use to measure the relationship between PMI and task accuracy: zero-shot task accuracy and accuracy gap.
On downstream datasets, we measure zero-shot accuracy for CLIP and VQA accuracy (as defined in \citep{agrawal2015vqa1}) for LMM evaluation tasks (evaluation details in Appendix~\ref{app:eval}).
We define \textit{accuracy gap} as the accuracy difference between inputs in the top and bottom 5\% of PMI values, representing the absolute accuracy degradation due to low PMI inputs.

\begin{figure}[b]
    \centering
    \begin{subfigure}[c]{0.45\textwidth}
        \includegraphics[width=1\linewidth]{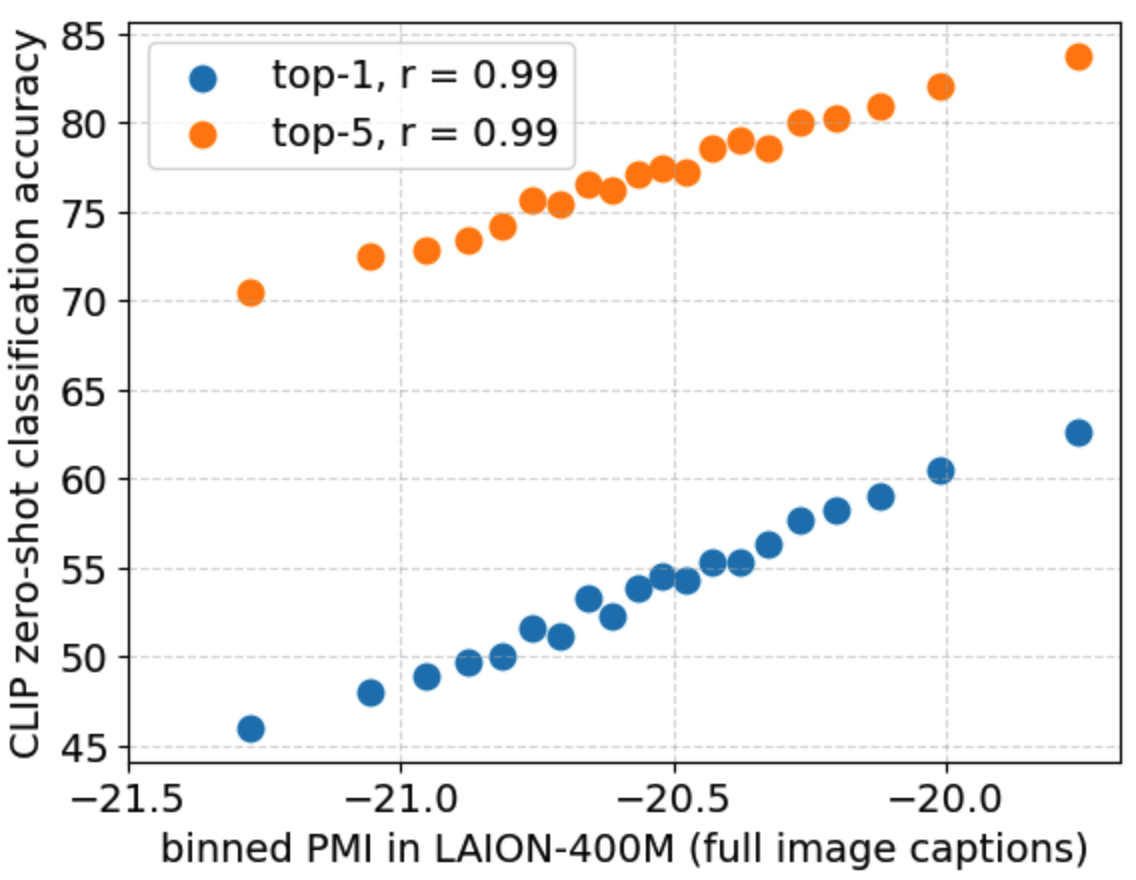}
        \vspace{-3ex}
        \subcaption{}
        \label{fig:synthetic-corr-allcaptions}
    \end{subfigure}
    \begin{subfigure}[c]{0.45\textwidth}
        \includegraphics[width=1\linewidth]{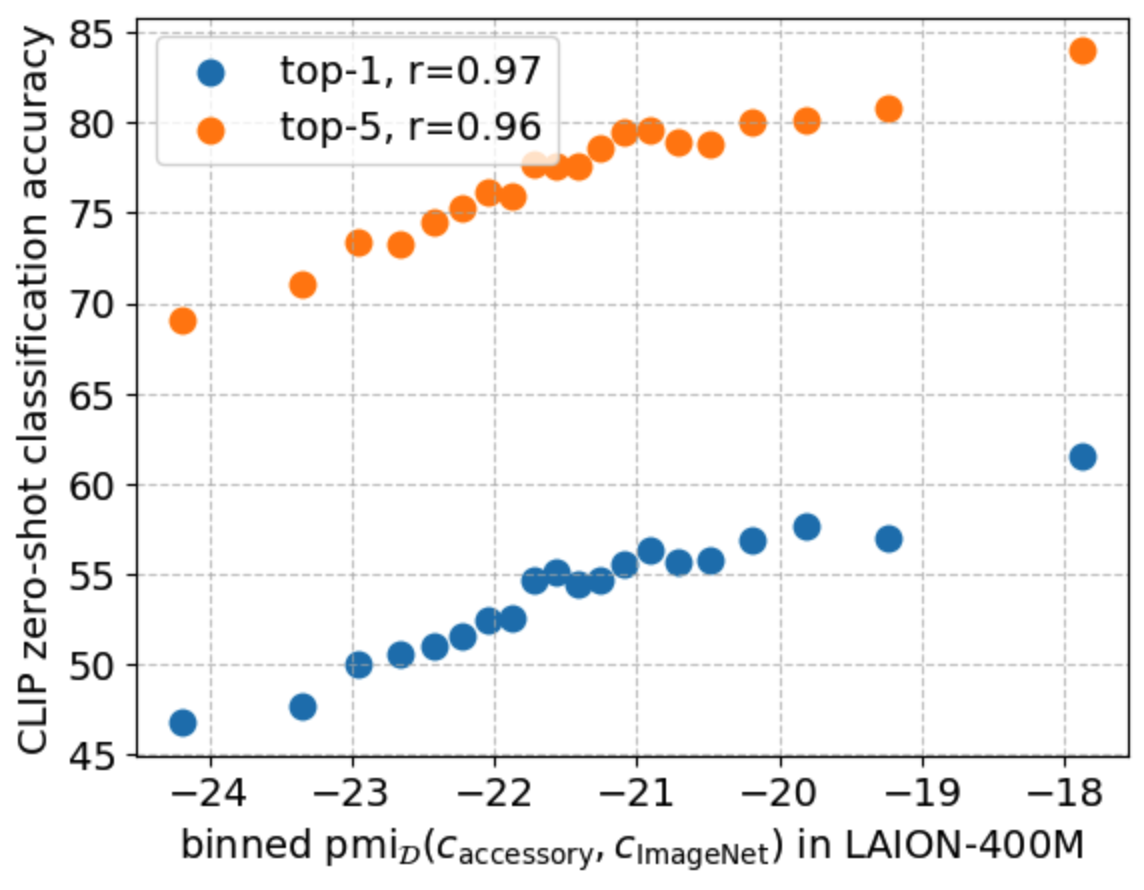}
        \vspace{-3ex}
        \subcaption{}
        \label{fig:synthetic-corr}
    \end{subfigure}
    \caption{\textbf{Strong correlation between concept PMI in pretraining data and CLIP zero-shot classification accuracy.} \textbf{(a)} We evaluate LAION-400M-trained CLIP on \synth, where each image depicts at least one concept, $\otherword$, in addition to the target ImageNet class $\imgnetword$. We observe a clear correlation between average PMI over all concepts in each image caption and CLIP zero-shot top-1 and top-5 accuracies, showing that pretraining concept co-occurrence strongly influences compositional generalization. \textbf{(b)} We observe a similar correlation between accuracy and the PMI of just the key concept pair, ($\otherword, \imgnetword)$.}
    \vspace{-3ex}
\end{figure}
\section{CLIP Zero-shot Classification Accuracy Correlates with Concept PMI}
\label{sec:synthetic}

We generate a dataset of synthetic images with a variety of concept pairs to investigate the relationship between PMI and CLIP zero-shot accuracy.

\paragraph{Task.}
We design input images for zero-shot classification that each feature at least two concepts, one of which is an ImageNet class, and evaluate CLIP's ability to predict the correct ImageNet class in the presence of the other concepts.
In order to control the set of concepts in each image, we generate synthetic images using text-to-image diffusion models.
\paragraph{Concept pairs.}

For this evaluation, we generate synthetic data using concept pairs that span the range of PMI in LAION-400M:
We first identify concept pairs (from the set of concepts extracted from LAION-400M) where only one of the two concepts is an ImageNet class.
To do so, we create a set of ImageNet class \textit{categories}, defined by the last word of each class name (e.g., $\texttt{king charles spaniel} \rightarrow \texttt{spaniel}$).
We select pairs where one of the two concepts matches an ImageNet category name and the other does not, including concept pairs that do not co-occur at all in LAION-400M as long as each individual concept is present.
Let such a pair be denoted $(\otherword, \imgnetword)$, where $\imgnetword$ is the ImageNet class word.
Finally, we filter the set of $\otherword$ to those that can be visualized in an image.
Further details can be found in Appendix~\ref{app:synthetic}.

\paragraph{GenPairs dataset construction.}
We construct our evaluation dataset, which we call \textit{\synth}, by generating a synthetic image for each concept pair extracted from LAION-400M where one of the two concepts is an ImageNet class, and define that ImageNet class as the ground truth label.
We subsample the set of concept pairs to obtain 200,000 pairs across the range of concept PMI.
We use Llama 3.1 8B Instruct \citep{grattafiori2024llama} to generate a realistic caption for an image that features each concept pair, and use these captions to prompt Flux.1-dev \citep{flux2024} to generate images (details in Appendix~\ref{app:synthetic}). We empirically find that Flux.1-dev produces realistic images even for low PMI pairs (see Figure~\ref{fig:synthetic-examples} for examples from \synth).

\paragraph{CLIP zero-shot classification accuracy correlates strongly with PMI of image caption concepts.}
We evaluate a CLIP ViT-B/32 pretrained on LAION-400M with \synth.
We calculate the average PMI over all valid concept pairs in each caption and show the correlation with classification accuracy in Figure~\ref{fig:synthetic-corr-allcaptions}.
We observe an $r=0.99$ correlation between PMI and CLIP zero-shot top-1 classification accuracy and an accuracy gap of 18\%, indicating that instead of generalizing to rarely seen concept combinations, CLIP's classification accuracy correlates predictably with the pretraining co-occurrence rate of the depicted concepts in the image.

\paragraph{CLIP zero-shot classification accuracy correlates strongly with PMI of key concept pair.}
Since the captions in \synth\ were explicitly generated to contain the key concept pair $(\otherword, \imgnetword)$, we analyze the correlation between zero-shot classification accuracy and the single PMI value for the key concept pair, $\pmifunc(\otherword, \imgnetword)$ (Figure~\ref{fig:synthetic-corr}).
We observe a $r=0.97$ correlation and an accuracy gap of 14\%.
This suggests that the PMI of the key concept pair alone is predictive of CLIP classification accuracy.

\begin{figure}
    \centering
    \begin{subfigure}[c]{0.45\textwidth}
        \includegraphics[width=1\linewidth]{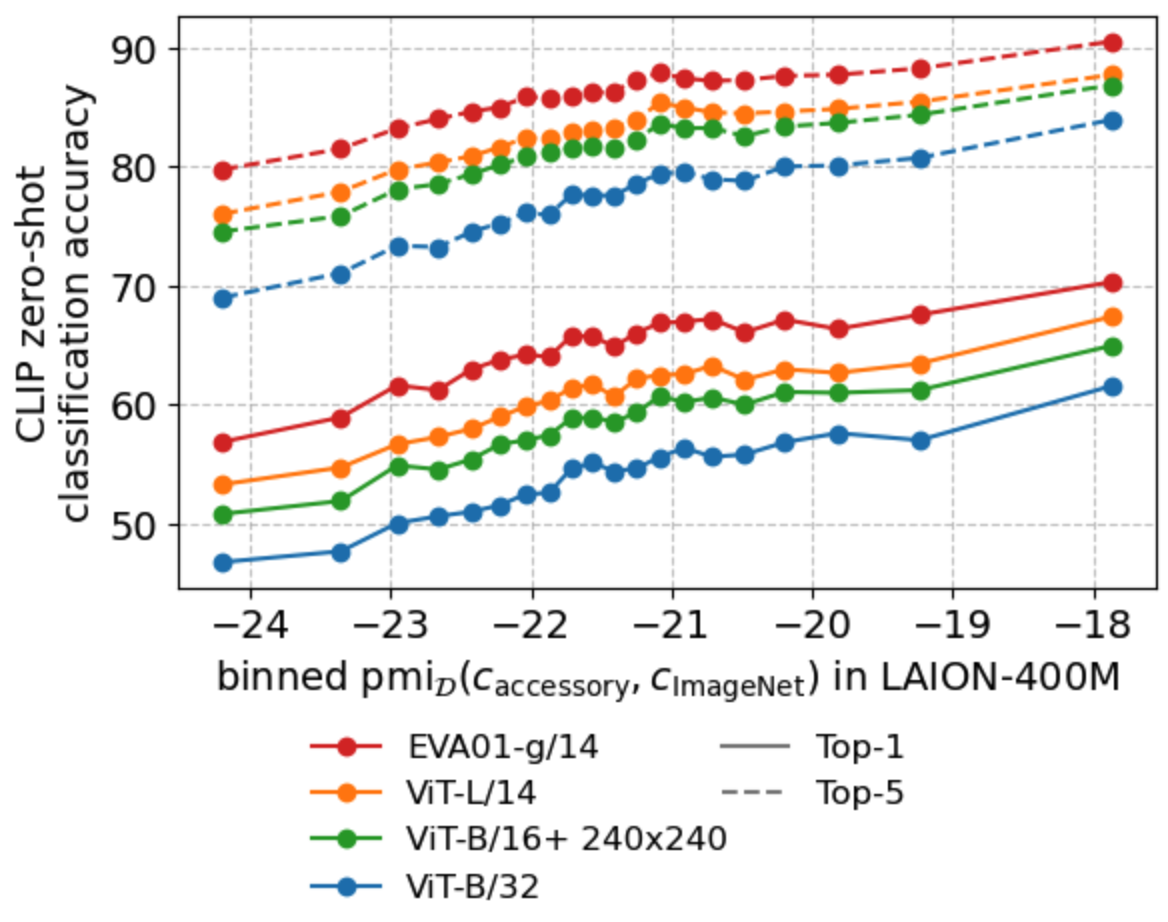}
        \vspace{-3ex}
        \subcaption{}
        \label{fig:scale>correlation}
    \end{subfigure}
    \begin{subfigure}[c]{0.45\textwidth}
        \includegraphics[width=1\linewidth]{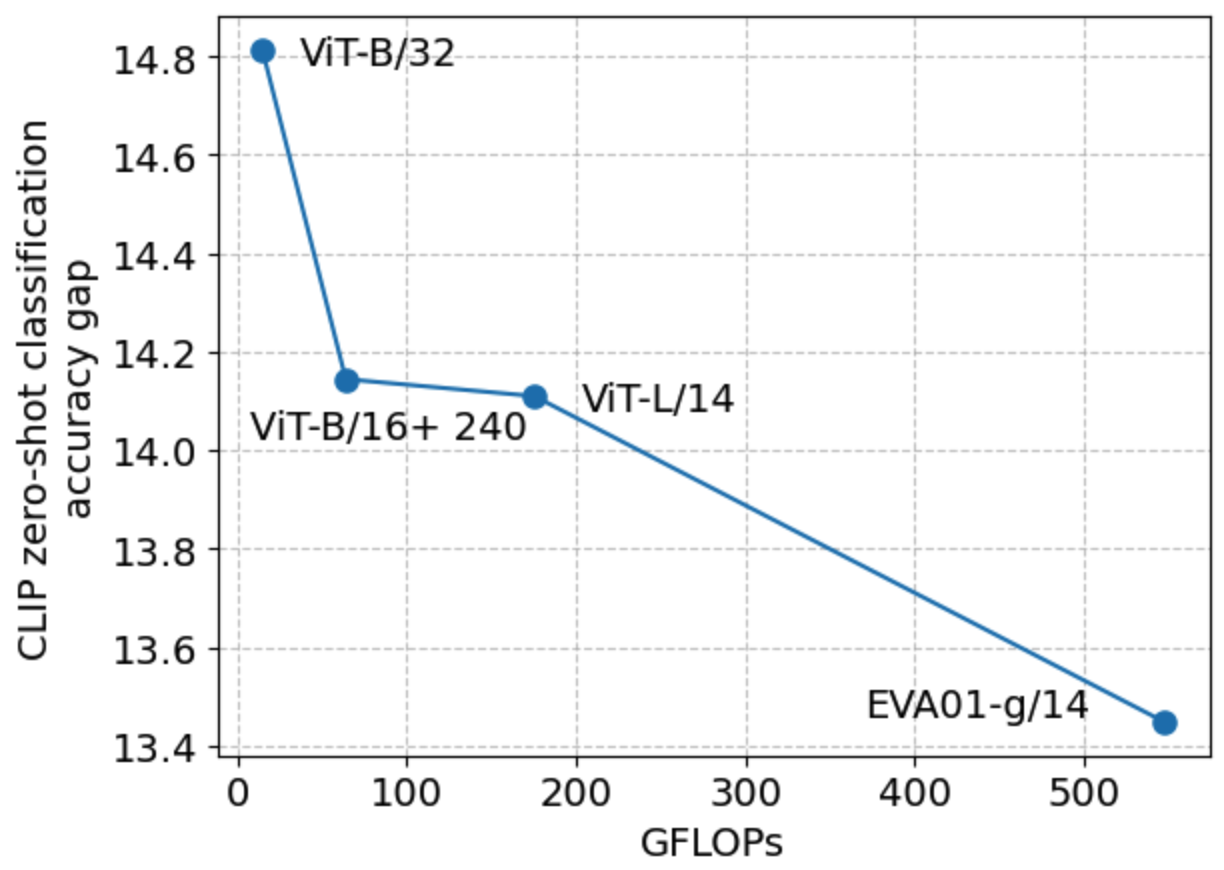}
        \vspace{-3ex}
        \subcaption{}
        \label{fig:scale>gap}
    \end{subfigure}
    \caption{\textbf{Accuracy gap improves slightly with model scale.} \textbf{(a)} In addition to ViT-B/32, we test 3 additional CLIP architectures pretrained with LAION-400M on \synth. \textbf{(b)} Accuracy gap on zero-shot classification decreases slightly with model scale.}
    \label{fig:scale}
\end{figure}

\paragraph{CLIP model scaling offers limited compositional generalization gains.}
Constructing a dataset with a balanced distribution over all concept combinations may reduce PMI-driven bias, but this approach becomes intractable with the number of possible combinations.
Instead, we investigate the impact of scaling the \textit{model} rather than the \textit{dataset}.
In addition to the ViT-B/32 baseline that is used for all CLIP experiments, we test 3 larger models all pretrained with LAION-400M (in increasing order of size: ViT-B/16+ 240, ViT-L/14, and EVA01-g/14) on \synth.
While the correlation persists across model scale (Figure~\ref{fig:scale>correlation}), Figure~\ref{fig:scale>gap} shows that the accuracy gap decreases slightly from 14.8\% in the smallest model, ViT-B/32, to 13.4\% in the largest model, EVA01-g/14 \citep{sun2023eva}.

\section{CLIP's zero-shot classification accuracy correlates with PMI in edited natural images}
\label{sec:pasting}

We observed in the previous section that while images often contain many concepts, a single key concept pair can be sufficient for analyzing the relationship between PMI and CLIP's zero-shot accuracy.
In this section, we use this insight to construct edits to natural images that introduce a particular concept pair to the image, affecting accuracy. 

\paragraph{Task.}
We test CLIP in the zero-shot classification setting on natural ImageNet images.
In order to quantify and vary the PMI of these natural images, we edit ImageNet validation set images by pasting a small image of another concept.
We then evaluate how CLIP's accuracy on edited images correlates with the PMI of the concept pair of the label and the pasted concept.

\paragraph{Concept pairs.}
We use the set of concept pairs defined in Section~\ref{sec:synthetic} where one of the two concepts is an ImageNet class.
Each pair can be denoted $(\otherword, \imgnetword)$, where $\imgnetword$ is the class of an ImageNet image.

\paragraph{ImageNet-Paste dataset construction.}
We construct our evaluation dataset, which we call \textit{\edited}, in two stages: first, we generate images of the set of accessory concepts using Flux.1-dev (details in Appendix~\ref{app:editing}).
For each ImageNet class $\imgnetword$, we sample a set of accessory concepts $\otherword$ across a range of PMIs $\pmifunc(\otherword,\imgnetword)$.
We scale the $\otherword$ image to be at most 10\% of the original image size, and paste it onto a randomly selected location (see Figure~\ref{fig:pasted-examples} for examples from \edited).

\paragraph{Strong correlation between PMI of edited natural images and CLIP zero-shot classification accuracy.}

\begin{figure}
    \centering
    \begin{subfigure}[c]{0.32\textwidth}
        \includegraphics[width=1\linewidth]{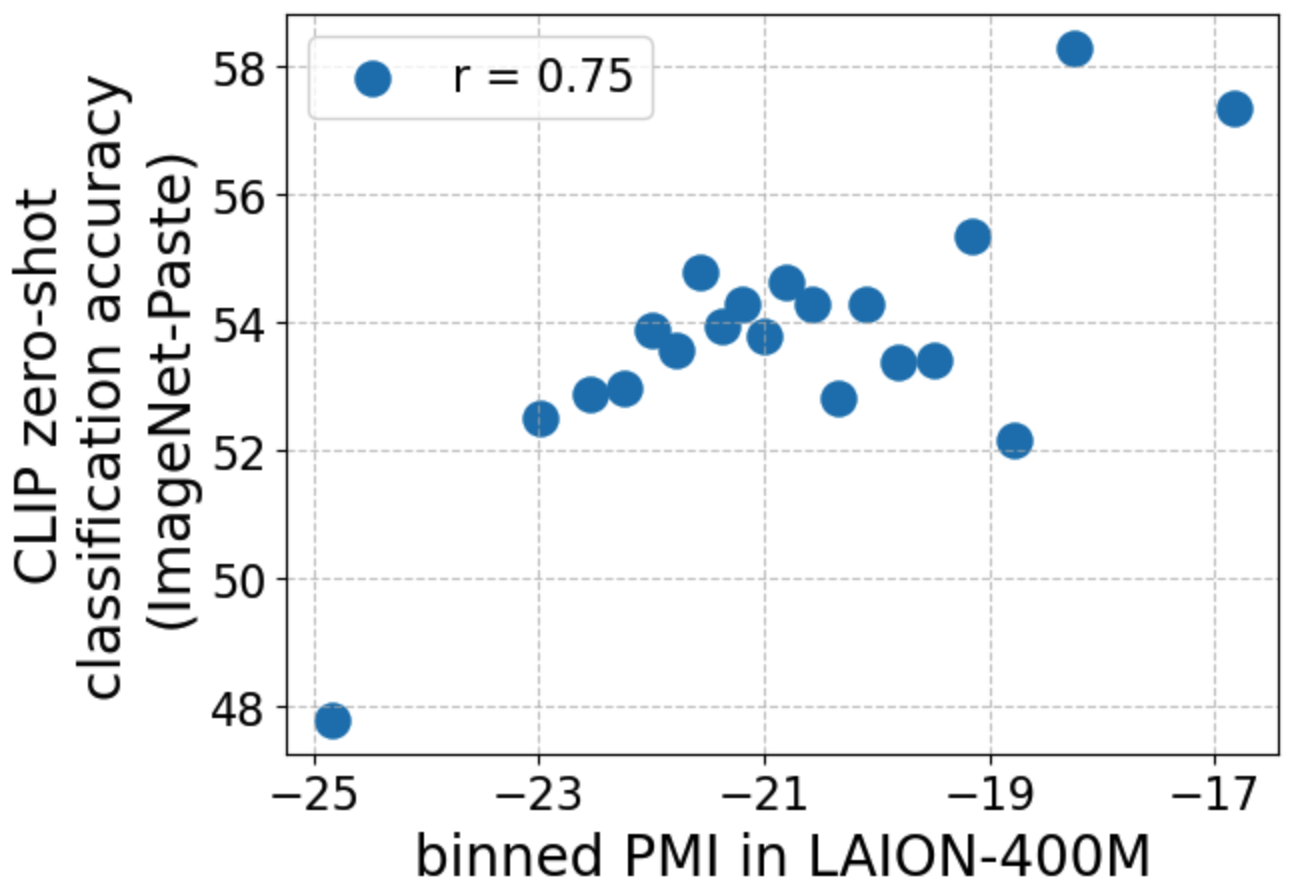}
        \subcaption{}
        \label{fig:pasted-corr}
    \end{subfigure}
    \begin{subfigure}[c]{0.32\textwidth}
        \includegraphics[width=0.95\linewidth]{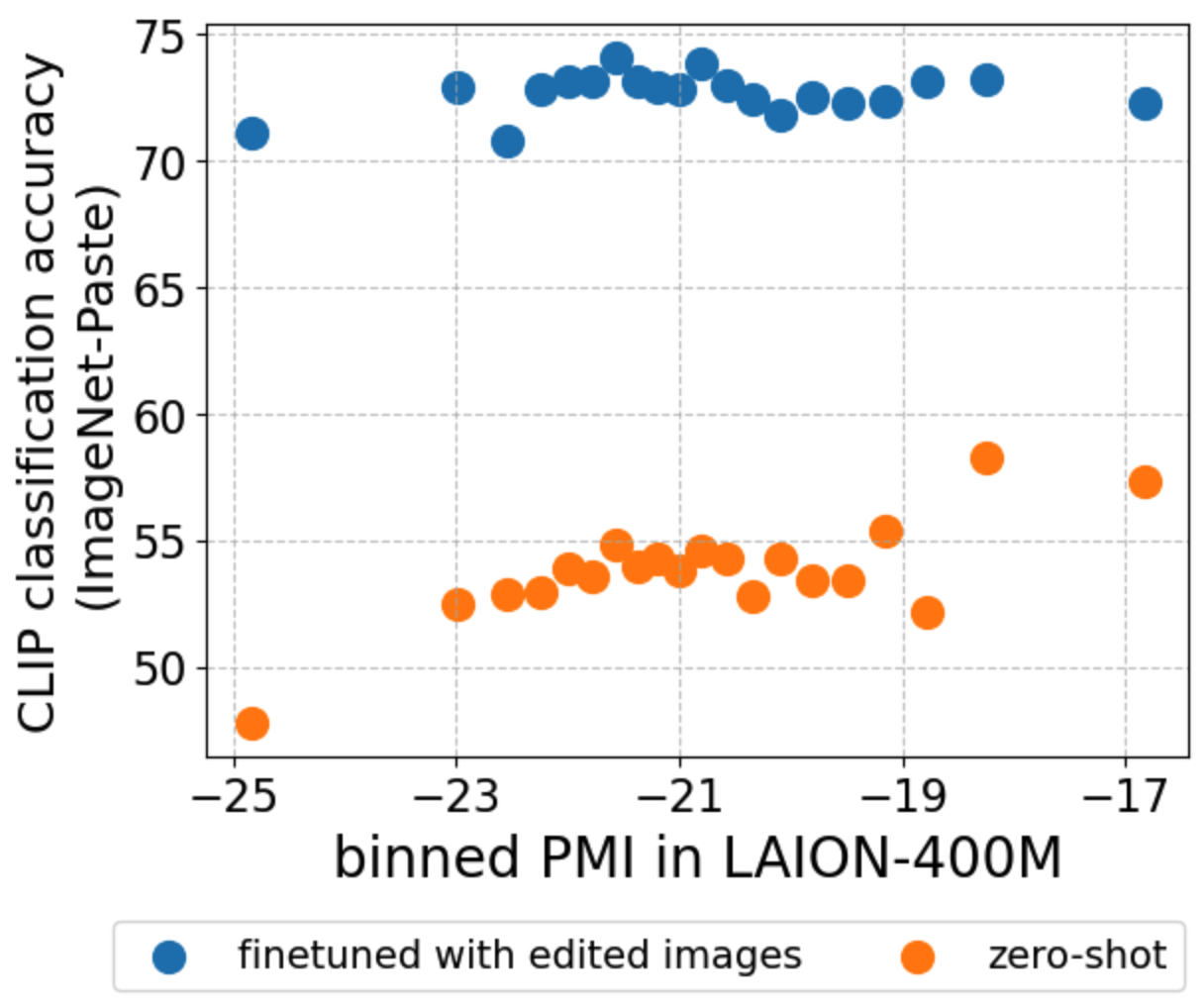}
        \subcaption{}
        \label{fig:pasted-ft-edited}
    \end{subfigure}
    \begin{subfigure}[c]{0.32\textwidth}
        \includegraphics[width=1\linewidth]{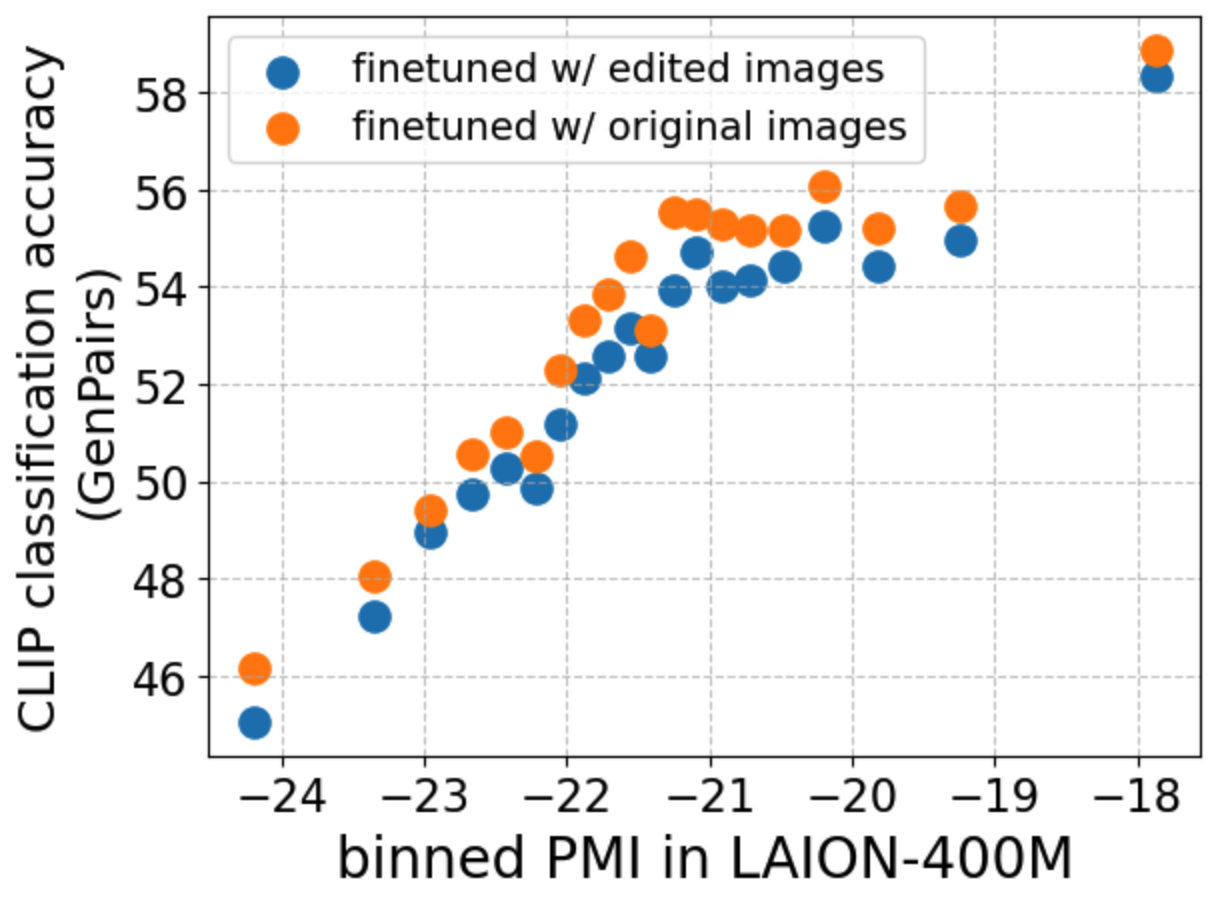}
        \subcaption{}
        \label{fig:pasted-ft-synthetic}
    \end{subfigure}
    \caption{\textbf{(a)} Editing natural (ImageNet validation) images by pasting an image of a concept with known PMI with the ImageNet class induces a correlation between zero-shot accuracy and PMI between the pasted and target class. In particular, pasting an image with a low PMI with the target class results in an accuracy drop. \textbf{(b)} Fine-tuning CLIP with edited images improves the overall accuracy and removes the correlation between PMI and accuracy on a held-out set of \edited. \textbf{(c)} However, fine-tuning CLIP on edited images does not remove the correlation between PMI and accuracy on our synthetic dataset, \synth, compared to fine-tuning on un-edited images.}
    \label{fig:pasted}
\end{figure}

Figure~\ref{fig:pasted-corr} shows the correlation between CLIP top-1 zero-shot classification accuracy and $\pmifunc(\otherword, \imgnetword)$ in \edited\ ($r=0.75$) with an accuracy gap of 10\%.
We note that the correlation we find is independent of the documented failures of CLIP models on classes that are poorly represented in pretraining, since the PMI metric is normalized by the individual concept frequencies.
Our concept-pair framework reveals a clear vulnerability in CLIP: even simple interventions, like pasting a concept with low PMI relative to the target class, can significantly degrade performance.

\paragraph{Fine-tuning CLIP with edited images substantially reduces the correlation when tested on edited images.}
We assess the impact of PMI-based image editing as an augmentation strategy for improved generalization to low PMI inputs.
We follow the procedure for generating image edits described above but implement it as an on-the-fly augmentation applied during a fine-tuning step to optimize CLIP embeddings to the ImageNet classification task.
Specifically, we perform end-to-end fine-tuning on the CLIP model for the ImageNet classification task with a linear projection layer (see Appendix~\ref{app:ft} for implementation details).
We evaluate CLIP fine-tuned with the augmentation on a held-out set of edited images and show our findings in Figure~\ref{fig:pasted-ft-edited}.
After fine-tuning with augmentation, the accuracy gap is reduced to 1\% compared to 10\% zero-shot. 

\paragraph{Accuracy gains from fine-tuning with edited images do not extend to other datasets.} 
To determine if fine-tuning with the image editing augmentation procedure can be a general strategy for improving accuracy for a range of PMIs, we test CLIP fine-tuned with the image editing augmentation on GenPairs (the PMI-controlled synthetic dataset described in Section~\ref{sec:synthetic}).
We find minimal difference in the PMI-accuracy correlation between fine-tuning with the image editing augmentation  compared to fine-tuning with unedited images (Figure~\ref{fig:pasted-ft-synthetic}), potentially suggesting that GenPairs and ImageNet-Paste test different aspects of the model that are both affected by concept co-occurrence.

\section{LMM Performance Correlates with Concept PMI}
\label{sec:lmm}

In this section, we extend our analysis from CLIP to large multimodal models (LMMs) that incorporate CLIP in their architecture.
We find that CLIP's failures to generalize to low PMI concept pairs affect downstream LMMs built on CLIP embeddings.

\paragraph{Task.}
We evaluate LMMs on the visual question-answering (VQA) task, which tests multimodal models' ability to understand visual inputs through open-ended natural language questions about images.
A VQA input example consists of an image and a natural language question about the image, and a set of possible ground truth answers produced by human annotators.
We identify concepts and calculate the PMI of each input VQA example by analyzing the question and answer text, and assess the LMM's ability to respond correctly to the question as a function of PMI.

\paragraph{Model.}
For this analysis, we use two variants of the LLaVA-1.5-7B model \citep{liu2023visual,liu2024improved}, a leading LMM built on CLIP image embeddings. 
The publicly available LLaVA-1.5-7B uses CLIP embeddings from OpenAI-trained CLIP ViT-L/14, which we denote LLaVA-1.5-OpenAI.
However, the OpenAI pretraining data is not publicly available.
In order to draw a direct comparison to CLIP pretraining data, we train our own version of LLaVA-1.5-7B that uses CLIP embeddings from LAION-400M-trained CLIP ViT-L/14, which we denote LLaVA-1.5-LAION.
We follow the visual instruction tuning procedure outlined in \citep{liu2024improved} to finetune a LLaVA-1.5 model with LAION-400M-trained CLIP as the vision backbone (details in Appendix~\ref{app:llava}).

\paragraph{Datasets.}
We evaluate on two standard visual question-answering benchmarks, VQAv2 \citep{goyal2017making} and TextVQA \citep{singh2019towards}.
VQAv2 is an open-ended VQA benchmark designed to test image understanding by targeting skills including object recognition, object counting, and relative locations.
VQAv2 includes a mix of yes/no and open-ended (not yes/no) questions, which we evaluate on separately.
TextVQA specifically focuses on question-answering with an optical character recognition (OCR) component.
We test on the validation split of VQAv2 (since the test split ground truth answers are not publicly available) and the test split of TextVQA.
We quantify performance using VQA accuracy as defined by the benchmarks.

\paragraph{Concept pairs.}
We adapt our PMI framework to the VQA setting by extracting concepts from the text of both the question and ground truth answer, as they both can contain information about the image (e.g., $\texttt{Q: who is wearing glasses? A: woman} \rightarrow \texttt{\{wear, glasses, woman\}}$).
We calculate PMI values for all concept pairs in each VQA example, then take the average for the final example-level PMI value.

\begin{figure}
    \centering
    \includegraphics[trim=0 31 0 0, width=0.85\linewidth,]{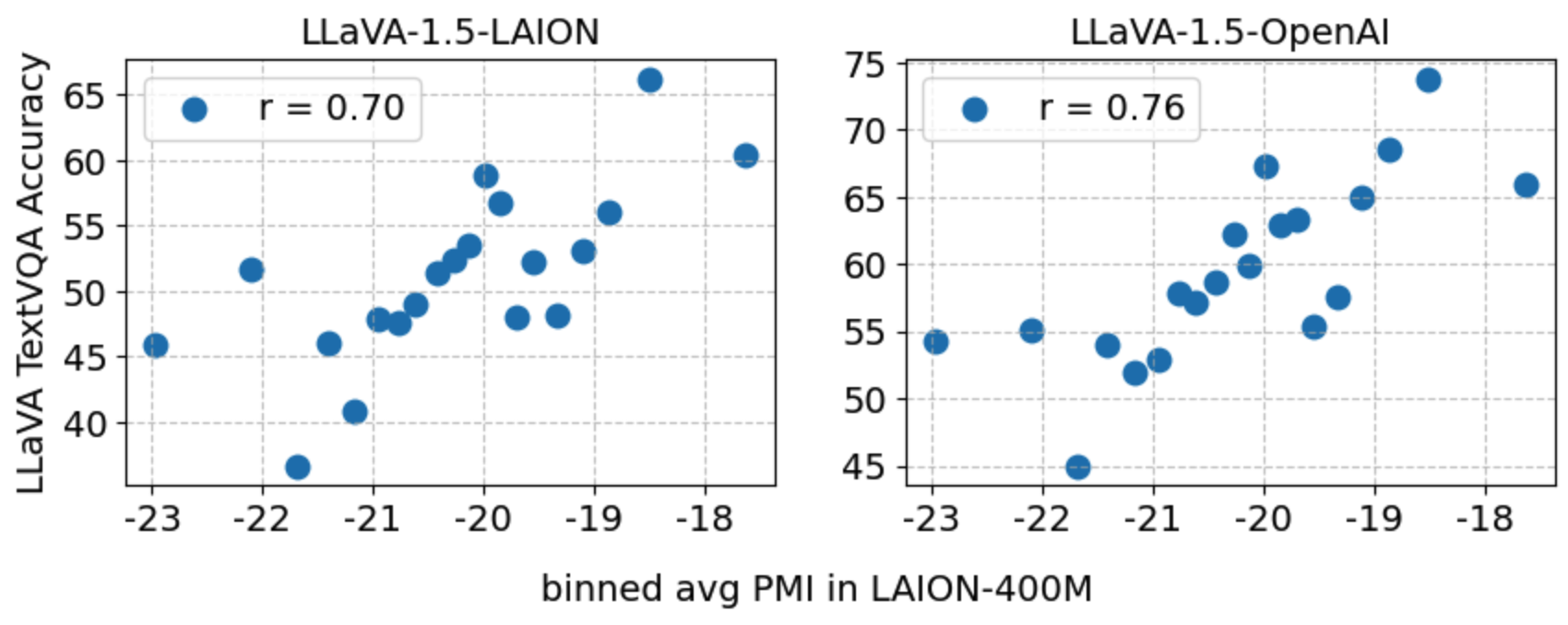}
    \includegraphics[width=0.87\linewidth]{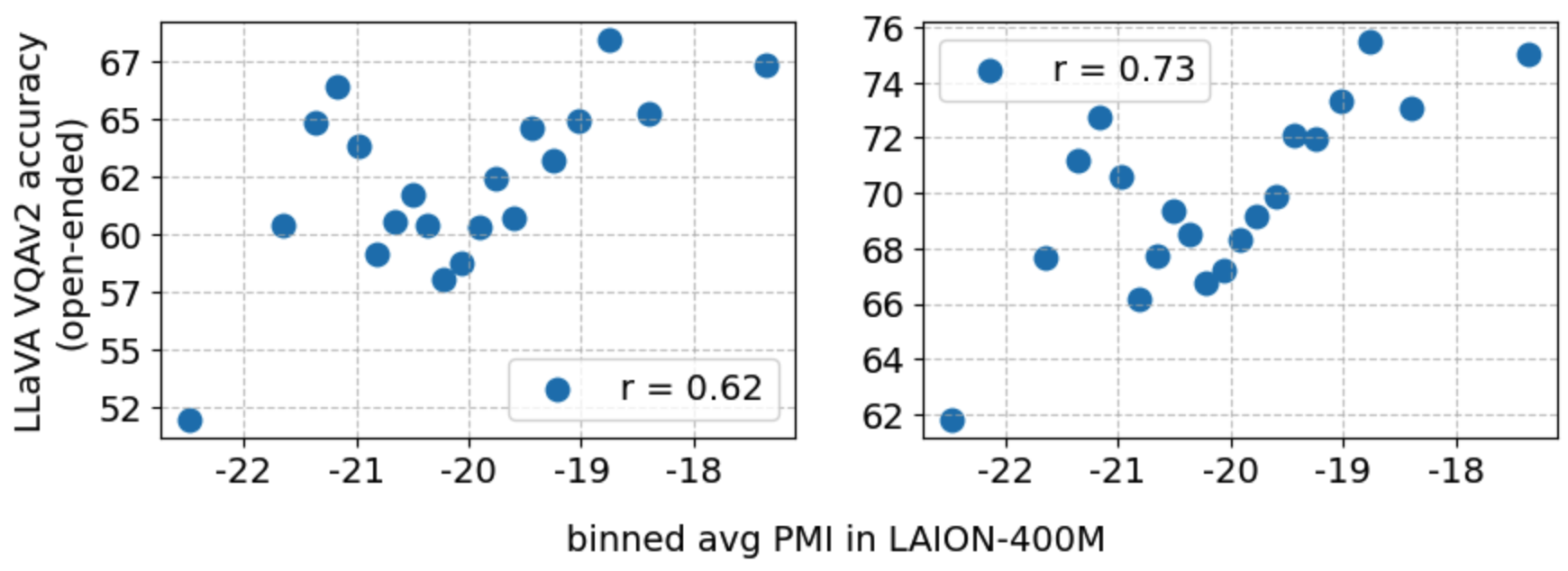}
    \caption{\textbf{Strong correlation between PMI in LAION-400M and LLaVA accuracy on VQA tasks .} We observe a strong correlation between LAION-400M LLaVA performance on both TextVQA \textbf{(top row)} and VQAv2 \textbf{(bottom row)}, where PMI for each input example is averaged across all concept pairs in the question and answer text. For VQAv2, we report performance on open-ended questions (all questions that require more than a `yes/no' response). LLaVA built on OpenAI CLIP \textbf{(right column)} also exhibits an almost identical correlation when using PMIs calculated with LAION-400M, despite OpenAI CLIP not being pretrained on LAION-400M.}
    \label{fig:vqa-corr}
\end{figure}

\begin{figure}
    \centering
    \begin{subfigure}[c]{0.48\textwidth}
        \includegraphics[width=1\linewidth]{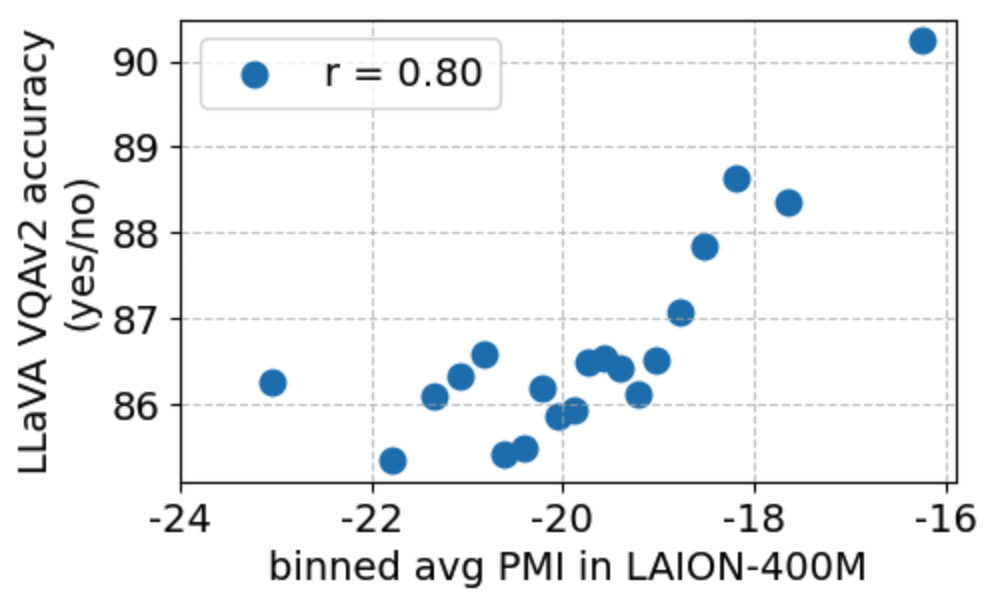}
        \subcaption{}
        \label{fig:vqav2-yesno-corr}
    \end{subfigure}
    \begin{subfigure}[c]{0.48\textwidth}
        \includegraphics[width=1\linewidth]{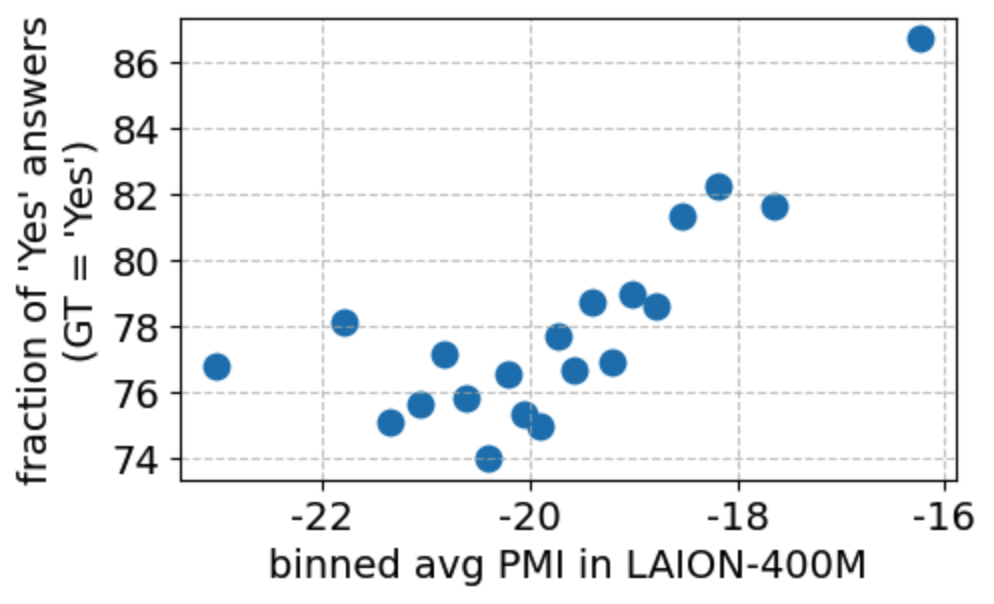}
        \subcaption{}
        \label{fig:vqav2-pyes}
    \end{subfigure}
    \caption{\textbf{(a) Strong correlation between PMI in LAION-400M and LLaVA accuracy on VQAv2 yes/no questions.} \textbf{(b) LMMs respond `yes' more often for higher average PMI inputs.} For VQAv2 questions with ground truth answer `yes', we find  the rate at which LAION-400M LLaVA correctly responds `yes' is highly correlated with the average PMI of concept pairs in the input question.}
    \label{fig:vqav2-Pyes}
\end{figure}

\paragraph{Strong correlation between PMI and LMM VQA accuracy.}
We evaluate LLaVA-1.5-LAION on TextVQA and VQAv2 and find a clear correlation with PMI. In particular:
\begin{itemize}
    \item \textbf{TextVQA:} We find a 15\% accuracy gap and a Pearson correlation coefficient of $r=0.70$  (Figure~\ref{fig:vqa-corr}, top left).
    \item \textbf{Open-ended VQAv2:} We observe a 15\% accuracy gap on open-ended VQAv2 examples (questions that require more than a binary yes/no response) and a correlation coefficient of $r=0.62$  (Figure~\ref{fig:vqa-corr}, top right).
    \item \textbf{Yes/No VQAv2:} We find a strong correlation for yes/no questions in VQAv2, with $r=0.80$ and a 4\% accuracy gap  (Figure~\ref{fig:vqav2-yesno-corr}).
\end{itemize}
We note that despite incorporating an LLM and performing visual instruction tuning steps that expose the full multimodal model to additional data, biases in the visual encoder still affect the performance of the downstream LMM.

\paragraph{Closed-source models exhibit an almost identical correlation between PMI and VQA accuracy.} We additionally test LLaVA-1.5-OpenAI on both benchmarks.
Despite the fact that OpenAI CLIP is trained on a closed-source dataset with few known properties, we observe that the correlation plots look almost identical to those of LLaVA-1.5-LAION other than an overall shift in accuracy.
We find a 10\% accuracy gap on TextVQA ($r=0.76$) and 13\% on VQAv2 ($r=0.73$) (Figure~\ref{fig:vqa-corr}, bottom row).
This suggests a shared long-tailed distribution of concept pairs in web-crawled datasets, and points to the possibility of performing data-centric analyses of closed-source model accuracy with open-source datasets.

\paragraph{LMMs bias toward responding `yes' with increasing PMI of concepts in the question.}
We additionally find that, for questions in the VQAv2 dataset with ground truth answer `yes', LLaVA-1.5-LAION's likelihood of correctly responding `yes' is  positively correlated ($r=0.77$) with the PMI of the concepts in the question (Figure~\ref{fig:vqav2-pyes}).
Here, we recalculate PMI with the concepts from the question only, not the answer (i.e., remove `yes' and `no' from the concept pairs), and find that the presence of high PMI concept pairs in the question is more likely to elicit a `yes' response.
Intuitively, this means that if concepts in the question co-occur often in the CLIP pretraining data, then the model is likely to respond `yes' regardless of the actual content in the image.
We note that this is an opposing effect to the correlation between PMI and accuracy described throughout this paper, as this bias worsens with higher PMI.
In this case, however, the overall accuracy improves despite this bias.

\begin{figure}
    \centering
    \begin{subfigure}[c]{0.32\textwidth}
    \includegraphics[width=1\linewidth]{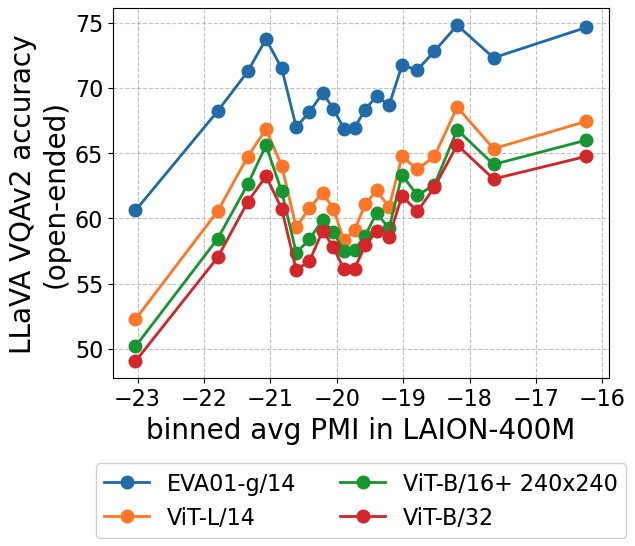}
        \label{fig:vqav2-openended-scale}
    \end{subfigure}
    \begin{subfigure}[c]{0.32\textwidth}
        \includegraphics[width=1\linewidth]{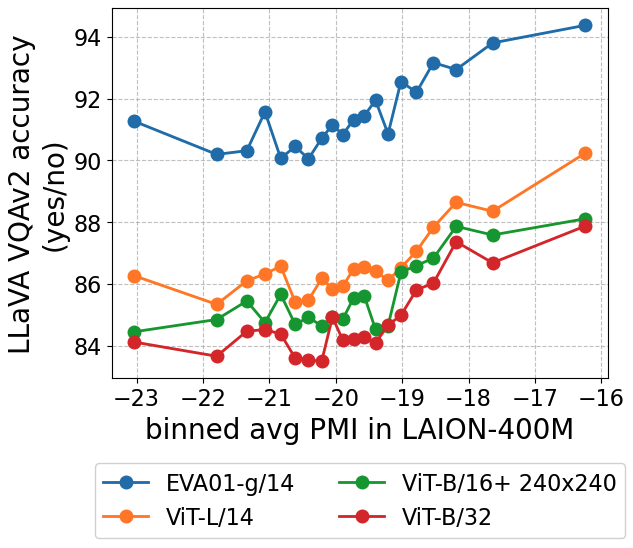}
        \label{fig:vqav2-yesno-scale}
    \end{subfigure}
    \begin{subfigure}[c]{0.31\textwidth}
        \includegraphics[width=1\linewidth]{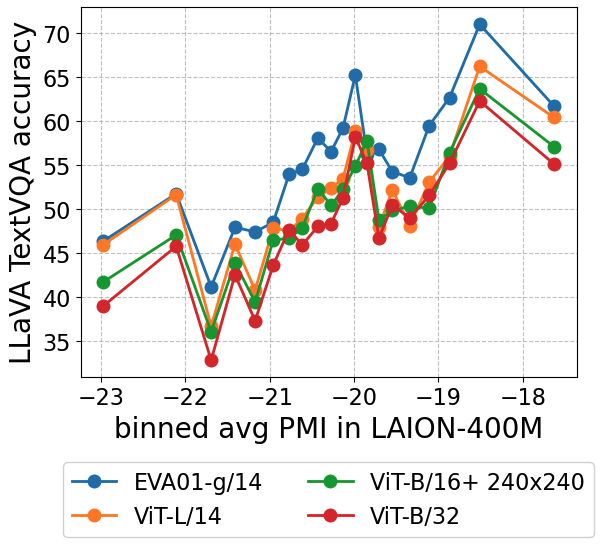}
        \label{fig:textvqa-scale}
    \end{subfigure}
    \begin{subfigure}[c]{0.32\textwidth}
        \includegraphics[width=1\linewidth]{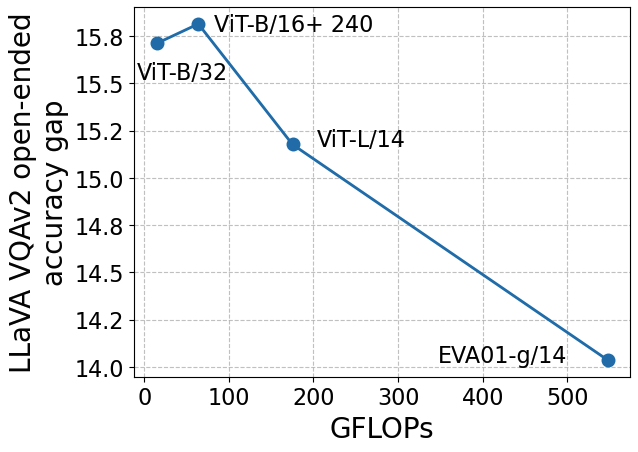}
        \label{fig:vqav2-openended-gap}
    \end{subfigure}
    \begin{subfigure}[c]{0.32\textwidth}
        \includegraphics[width=1\linewidth]{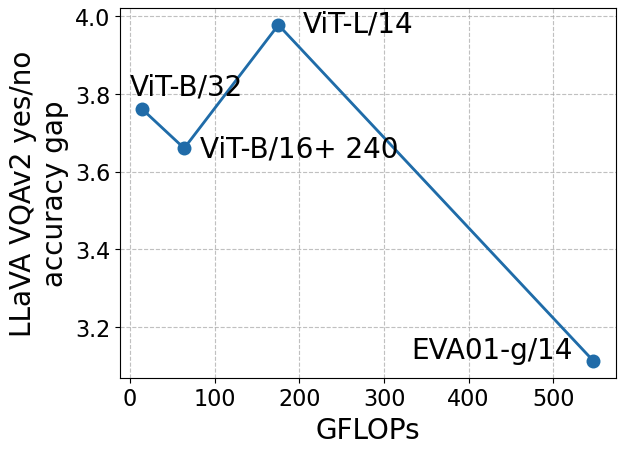}
        \label{fig:vqav2-yesno-gap}
    \end{subfigure}
    \begin{subfigure}[c]{0.32\textwidth}
        \includegraphics[width=1\linewidth]{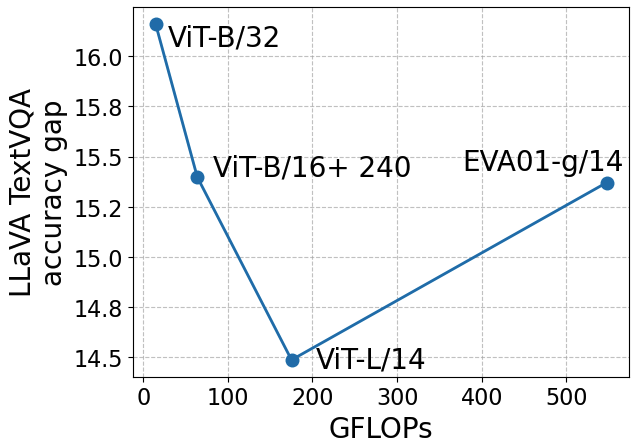}
        \label{fig:textvqa-gap}
    \end{subfigure}
    \caption{\textbf{CLIP model scale does not consistently improve generalization to low PMI inputs in LMMs.} \textbf{(top row)} In addition to the default CLIP ViT-L/14, we train LLaVA-1.5-7B models based on 3 additional CLIP architectures and test them on VQAv2 and TextVQA. \textbf{(bottom row)} CLIP model scale is not consistently predictive of accuracy gap across tasks.}
    \label{fig:vqa-scale}
\end{figure}

\paragraph{Scaling CLIP in the LMM context.}
We extend our finding in Section~\ref{sec:synthetic} regarding the impact of CLIP model scale on compositional generalization to LMMs.
Specifically, we train LLaVA-1.5-7B models with the 4 LAION-400M-pretrained CLIP backbones from Figure~\ref{fig:scale}: ViT-B/32, ViT-B/16+ 240, ViT-L/14, and EVA01-g/14 (in increasing order of size).
We note that the default CLIP model size for LLaVA-1.5-7B is ViT-L/14, which was used for all other LLaVA experiments in this section.
We observe that correlation between PMI and VQA accuracy persists across scales (Figure~\ref{fig:vqa-scale}, top row) and the relationship between accuracy gap and model size varies between tasks (Figure~\ref{fig:vqa-scale}, bottom row); however, the largest model (EVA01-g/14) consistently produces a smaller accuracy gap compared to the smallest (ViT-B/32) ($15.8 \rightarrow 14.0$ for VQAv2 (open-ended), $3.8 \rightarrow 3.1$ for VQAv2 (yes/no), $16.1 \rightarrow 15.3$ for TextVQA).
These results suggest that, in the context of CLIP-based LMMs, scaling CLIP alone may not robustly produce gains in compositional generalization.

\section{Related Work and Discussion}
\label{sec:discussion}
\paragraph{Spurious correlations.}
Machine learning models tend to learn spurious correlations present in the training dataset \citep{beery2018recognition,zech2018variable,sagawa2020group,geirhos2020shortcut}.
Recent work \citep{wang2024sober} demonstrates that even web-scale pretrained models like CLIP are not immune to spurious correlations, such as wildlife in likely vs. unlikely environments.
Our work supports these results, while quantifying the relationship between concept co-occurrence in the training data and downstream accuracy. Data reweighting interventions such as Group DRO~\citep{sagawa2020group} required group labels, while our PMI metric could enable new data-centric algorithms for compositional generalization.

\paragraph{Compositional generalization in multimodal models.}
Compositional understanding and generalization to novel combinations of concepts is essential for effective models.
However, it has proven a challenging ability to learn since early vision-language models \citep{johnson2017clevr,bogin2021covr}.
Recent works evaluate CLIP's performance on concept combinations unseen during pretraining \citep{abbasi2024deciphering,wiedemer2025pretraining}, where \citep{wiedemer2025pretraining} shows performance scales as a function of individual concept frequencies in pretraining.
In this work, we go a step further to quantify concept co-occurrence in pretraining and find a strong correlation with CLIP accuracy.
Many compositionality benchmarks exist for LMMs \citep{lewis2022does,yuksekgonul2022and,thrush2022winoground,hsieh2023sugarcrepe}, but the role of pretraining data in determining benchmark accuracy is unclear.
We extend our result to CLIP-based LMMs and show that a clear correlation persists between LMM VQA accuracy and CLIP pretraining concept co-occurrence.

\paragraph{LMMs and evaluation.} 
Modern LMMs are primarily built by combining embeddings from a frozen visual encoder, most commonly CLIP, with a large language model \citep{li2023blip,liu2023visual,awadalla2023openflamingo,deitke2024molmo,liu2024improved,tong2024cambrian}.
As such, failures of the visual encoder can have a direct impact on the efficacy of the downstream LMM \citep{tong2024eyes}.
Existing evaluations of LMMs test for a wide variety of capabilities \citep{singh2019towards, goyal2017making, hua2024mmcomposition, tong2024cambrian, ma2023crepe, hsieh2023sugarcrepe, thrush2022winoground, liu2024mmbench}, but the connection between task performance and the pretraining data distribution of the visual encoder has not been explored.
We fill this important gap by demonstrating that the relationship between concept PMI in pretraining and CLIP performance on those concepts extends to CLIP-based LMMs, underscoring the importance of a robust visual encoder.

\paragraph{Limitations.}
We acknowledge several limitations in our methods.
Our concept extraction pipeline assumes that concepts present in text captions are an accurate proxy for visual concepts, but validating this is difficult without a comparison against human-annotated images.
In addition, though we empirically validate our data generation process, both our CLIP experiments rely in some form on text-to-image diffusion models, which are themselves trained on a data distribution and may generalize poorly to uncommon prompts or concepts.

\section{Conclusion}
Our study reveals that CLIP and LMMs built on CLIP are highly sensitive to the co-occurrence statistics of concept pairs in their pretraining data. 
This leads to a strong correlation between PMI of inputs and task accuracy as well as sizable observed accuracy gaps between high and low PMI concept pairs across multiple datasets and tasks, showing that these models struggle to disentangle individual concepts and generalize to new combinations.
Closing this gap will require new methods that promote robust compositional generalization without relying on combinatorially large datasets.

\begin{ack}
We thank Oliver Liu for helpful discussions, Chad Popik for help with graphics, and the Scientific Computing Core at the Flatiron Institute, a division of the Simons Foundation, for computing resources and support.
\end{ack}

\bibliography{references}


\newpage
\appendix

\section{Implementation Details}

\subsection{Additional Details on Concept Extraction and PMI Calculation}
\label{app:concepts}
We use the $\texttt{nltk}$ package to clean, lemmatize, and perform part-of-speech tagging for the LAION-400M captions.
To treat pairs with $\pairprob = 0$ or $\singleprob = 0$, we calculate all PMI frequency ratios with Laplace smoothing with a smoothing factor of $\alpha = 1$ for $\pairprob$ and $\alpha = \text{1e4}$ for $\singleprob$.

\subsection{Synthetic Data Generation}
\label{app:synthetic}
After obtaining the set of $(\otherword, \imgnetword)$ concept pairs, we want to retain a set of $\otherword$ that are visualizable English words. To do this, we first perform basic filtering: we remove $\otherword$ with numeric digits or that are not in a WordNet synset (as a proxy for non-English words or non-words), then perform POS tagging and keep only nouns and adjectives that are not the words $\texttt{photo}$ or $\texttt{image}$.
Finally, we prompt Llama 3.1 8B Instruct to distinguish $\otherword$ that are ``visualizable'' in order to filter out words like ``new'' or ``success'' that are difficult to incorporate into an image.
We reproduce the prompt in Block~\ref{lst:visualizable_prompt}.

After filtering, we generate a one-sentence image caption for each concept pair using Llama 3.1 8B Instruct and the prompt in Block~\ref{lst:llm_prompt}.
We then prompt the text-to-image model Flux.1-dev with the captions produced in the previous step.
The hyperparameters we used for Llama and Flux.1-dev are detailed in Tables~\ref{tbl:llm-params} and~\ref{tbl:flux-params}.
We empirically find that these hyperparameters produce the most realistic captions and images for our purposes.
We use HuggingFace implementations of both models.

We use the OpenCLIP implementation of all models \citep{cherti2023reproducible} and note that EVA01-g/14 is from the EVA-CLIP family of models \citep{sun2023eva}.

\begin{figure}
    \centering
    \includegraphics[width=0.49\linewidth]{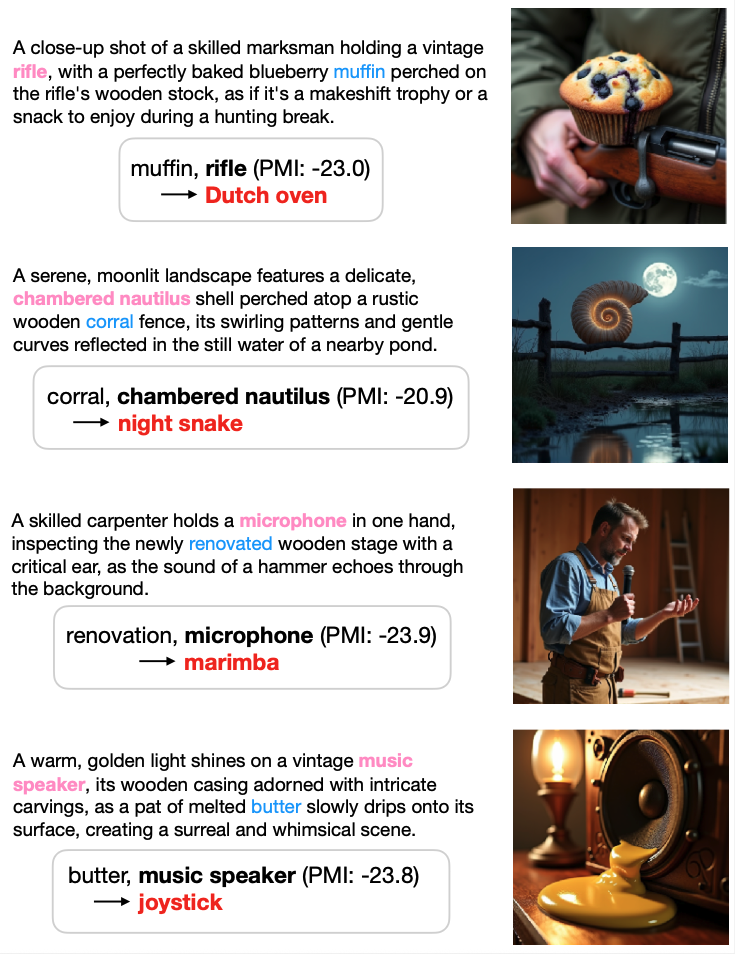}
    \includegraphics[width=0.49\linewidth]{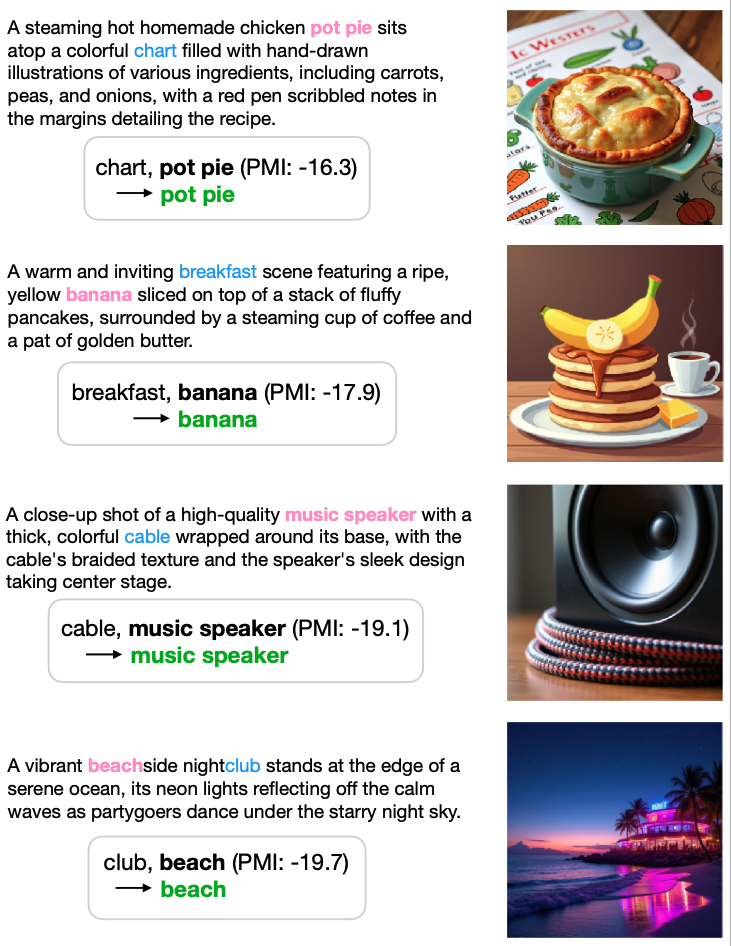}
    \caption{\textbf{Examples from \synth} (\textbf{left}: low PMI, \textbf{right}: high PMI). We use Meta's Llama 3.1 8B Instruct to generate captions for images incorporating the concept pairs ($\otherword$ in pink, $\imgnetword$ in bold/blue). We prompt Flux.1-dev with the captions to produce the images shown. Finally, CLIP's zero-shot prediction on each image is shown (red: incorrect, green: correct).}
    \label{fig:synthetic-examples}
\end{figure}

\begin{lstlisting}[style=promptblock, caption={Prompt used for $\otherword$ ``visualizability'' filtering with Llama. $\texttt{\{c\}}$ is replaced with the concept word.}, label={lst:visualizable_prompt}]
You will be provided with some examples of questions and answers determining whether a word is easily visualizable, followed by a question for you to solve. An easily visualizable word is a concrete thing or adjective that describes the subject of an image. Abstract concepts that can be represented by concrete objects/images are NOT easily visualizable. When in doubt, answer no. Please think aloud step-by-step and conclude your answer with the phrase "The answer is X.". You must use exactly this phrase, otherwise we will be unable to use your answer.

## Examples

Q: Is temperament easily visualizable?
A: Let's think step by step. Temperament is a property of a person/animal, so the subject of the image would be that person/animal and not "temperament". The answer is no.

Q: Is sb easily visualizable?
A: Let's think step by step. Sb is not a word and is thus not visualizable. The answer is no.

Q: Is fertilizer easily visualizable?
A: Let's think step by step. Fertilizer is a concrete object and can be visualized by, e.g., a bag of fertilizer. The answer is yes.

Q: Is impressionism easily visualizable?
A: Let's think step by step. Impressionism is an art style so images can be rendered in an impressionist style. The answer is yes.

Q: Is browsing easily visualizable?
A: Let's think step by step. Browsing is an action, and actions are not directly visualizable in a static image. The answer is no.

Q: Is success easily visualizable?
A: Let's think step by step. Success is an abstract concept. It could be represented by a trophy or other concrete object, but then that object would be the subject of the image, so it is not directly visualizable. The answer is no.

Q: Is helen easily visualizable?
A: Let's think step by step. Helen is a proper noun, likely referring to a person named Helen, but this would be impossible to know without a text description. Helen is thus not visualizable. The answer is no.

## Your Question
Q: Is {c} easily visualizable?
A: Let's think step by step.
\end{lstlisting}

\begin{lstlisting}[style=promptblock, caption={Prompt used for image caption generation with Llama. $\texttt{\{c1\}}$ and $\texttt{\{c2\}}$ are replaced with the concepts in the concept pair.}, label={lst:llm_prompt}]
Please write a single sentence that could describe an image that contains the words `{c1}' and `{c2}'. Make sure both {c1} and {c2} are the focus of the image.
\end{lstlisting}

\begin{table}[]
    \centering
    \begin{tabular}{lc}
        \toprule
        parameter & value \\
        \midrule
        temperature & 0.1 \\
        minp & 0.05 \\
        max new tokens & 50  \\
        \bottomrule
    \end{tabular}
    \caption{Llama hyperparameters for visualizability filtering and caption generation.}
    \label{tbl:llm-params}
\end{table}

\begin{table}[]
    \centering
    \begin{tabular}{lc}
        \toprule
        parameter & value \\
        \midrule
        output size (px) & $512 \times 512$ \\
        guidance scale & 5.0 \\
        inference steps & 28  \\
        \bottomrule
    \end{tabular}
    \caption{Flux.1-dev hyperparameters for generating the images for Section~\ref{sec:synthetic} as well as the pasted images for Section~\ref{sec:pasting}.}
    \label{tbl:flux-params}
\end{table}

\subsection{Natural Image Editing}
\label{app:editing}
\begin{figure}
    \centering
    \includegraphics[width=1\linewidth]{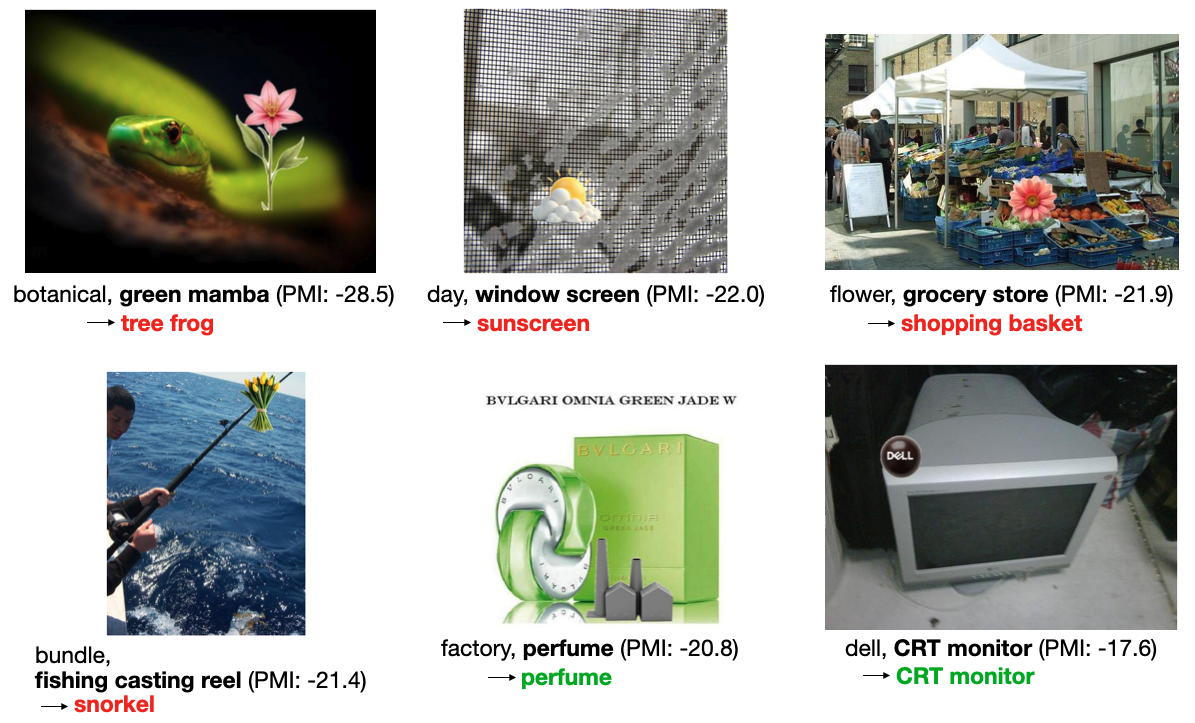}
    \caption{\textbf{Examples from our edited natural images dataset described in Section~\ref{sec:pasting}.} We prompt Flux.1-dev to generate images of a set of $\otherword$ accessory concepts, then paste onto an ImageNet validation set image of class $\imgnetword$. The concept pair is shown under each image, with $\imgnetword$ shown in bold, as well as CLIP's zero-shot prediction on each edited image (red: incorrect, green: correct).}
    \label{fig:pasted-examples}
\end{figure}

We generate an image of each $\otherword$ by prompting Flux.1-dev with the simple phrase ``$\texttt{a \{}\otherword \texttt{\} in the center of a white background}$''.
As these accessory images will be pasted on top of ImageNet images, we replace the pasted images' white background with transparent pixels to emulate the concept occurring in the image ``naturally''.
To do so, we generate an object mask with the Segment Anything \citep{kirillov2023segment} object segmentation model and assign masked background pixels to fully transparent using the RGBA format.
All image manipulations were done with the $\texttt{PIL}$ Python package.
Examples of edited images are shown in Figure~\ref{fig:pasted-examples}.

\subsection{CLIP Fine-tuning}
\label{app:ft}

We fine-tune CLIP for the ImageNet classification task starting with a linear layer initialized with CLIP's zero-shot ImageNet classification weights. We follow the WiSE-FT fine-tuning recipe \citep{wortsman2021robust} to fine-tune end-to-end with learning rate 3e-5 with 500 steps of linear warmup, weight decay 0.1, batch size 512, and train for 10 epochs. We train with the ImageNet training split with or without the image editing augmentation described in Section~\ref{sec:pasting}.

\subsection{LLaVA Experiments}
\label{app:llava}
\paragraph{LLaVA fine-tuning.} We fine-tune our own LLaVA-1.5-7B with LAION-400M CLIP ViT-L/14. We follow the LLaVA-1.5 visual instruction tuning recipe and first pretrain the vision-language connector with CLIP and LLM both frozen. We keep the same hyperparameters as they recommend but empirically find that a lower maximum learning rate of 1e-4 is more effective. After the pretraining step, we fine-tune both the connector and the LLM using the published LLaVA-1.5 visual instruction tuning dataset with the suggested hyperparameters.

\paragraph{Processing the VQA datasets.} As TextVQA and VQAv2 answers come from real human responses with some variance, we define the ``ground truth answer'' as the mode of the collected human responses.
We tokenize, lemmatize, and remove stopwords (we do not remove `yes' and `no' since some examples are yes/no questions) from each question-answer pair in the VQA datasets to obtain the set of concepts and concept pairs for each example.

\subsection{Evaluation Details}
\label{app:eval}
We follow the OpenCLIP implementation \citep{cherti2023reproducible,ilharco_gabriel_2021_5143773} of the original CLIP \citep{radford2021clip} work's zero-shot classification recipe for all zero-shot CLIP evaluations on GenPairs and ImageNet-Paste. We follow the directions in the official LLaVA repository\footnote{\texttt{https://github.com/haotian-liu/LLaVA/blob/main/docs/Evaluation.md}} to evaluate LLaVA-1.5-LAION and LLaVA-1.5-OpenAI on TextVQA and VQAv2.

\section{Compute Requirements}
\label{app:compute}
We ran experiments on a combination of NVIDIA A100 and H100 GPUs. Non-trivial compute was needed for:
\begin{itemize}
    \item generating captions with Llama: 0.2s/caption on a single GPU
    \item generating images with Flux.1-dev: 4.7s/image on a single GPU
    \item end-to-end fine-tuning of CLIP: 7 hrs on 4 A100s
    \item training vision-language connector for LLaVA-1.5: 40 mins on 4 H100s
    \item visual instruction tuning for LLaVA-1.5: 8.5 hrs on 8 H100s
\end{itemize}
We estimate the total compute to be $\sim 1$ month of GPU time, including preliminary or failed experiments.

\section{Licenses}
\label{app:licenses}
\begin{itemize}
    \item ImageNet \citep{deng2009imagenet} is licensed under BSD 3-Clause License.
    \item LAION \citep{schuhmann2021laion} is licensed under MIT License.
    \item TextVQA \citep{singh2019towards} is licensed under CC BY 4.0 License.
    \item VQAv2 \citep{singh2019towards} is licensed under CC BY 4.0 License.
    \item Flux.1-dev \citep{flux2024} is under a Non-Commercial License.
    \item LLaVA \citep{liu2023visual} is licensed under the Apache License 2.0.
    \item CLIP \citep{radford2021clip} and OpenCLIP \citep{cherti2023reproducible} are licensed under MIT License.
\end{itemize}

\end{document}